\documentclass[a4paper,twoside]{article}

\usepackage{epsfig}
\usepackage{calc}
\usepackage{amssymb}
\usepackage{amstext}
\usepackage{amsmath}
\usepackage{amsthm}
\usepackage{multicol}
\usepackage{pslatex}
\usepackage{apalike}

\usepackage[caption=false,font=footnotesize,position=top,singlelinecheck=off,justification=raggedright]{subfig}
\usepackage{siunitx}
\usepackage{SCITEPRESS}     

\begin{document}

\title{The influence of labeling techniques in classifying human manipulation movement of different speed.}

\author{\authorname{Sadique Adnan Siddiqui\sup{1}, Lisa Gutzeit\sup{1} and Frank Kirchner\sup{1,2}}
\affiliation{\sup{1}Robotics Research Group, University of Bremen, Bremen, Germany}
\affiliation{\sup{2} Robotics Inovation Center, DFKI, Bremen, Germany}
\email{\{siddiqui, Lisa.Gutzeit\}@uni-bremen.de, Frank.Kirchner@dfki.de}
}



\abstract{
Human action recognition aims to understand and identify different human behaviors and designate appropriate labels for each movement's action. In this work, we investigate the influence of labeling methods on the classification of human movements on data recorded using a marker-based motion capture system. The dataset is labeled using two different approaches, one based on video data of the movements, the other based on the movement trajectories recorded using the motion capture system.
The data was recorded from one participant performing a stacking scenario comprising simple arm movements at three different speeds (slow, normal, fast). Machine learning algorithms that include k-Nearest Neighbor, Random Forest, Extreme Gradient Boosting classifier, Convolutional Neural networks (CNN), Long Short-Term Memory networks (LSTM), and a combination of CNN-LSTM networks are compared on their performance in recognition of these arm movements. The models were trained on actions performed on slow and normal speed movements segments and generalized on actions consisting of fast-paced human movement. It was observed that all the models trained on normal-paced data labeled using trajectories have almost 20\% improvement in accuracy on test data in comparison to the models trained on data labeled using videos of the performed experiments. 
}

\onecolumn \maketitle \normalsize \setcounter{footnote}{0} \vfill

\section{\uppercase{Introduction}}
Recognition of human actions is an active research area utilizing both vision and non-vision based modalities. Machine Learning and Deep Learning algorithms have shown promising results in the identification and understanding of human behaviors, which is important to improve the collaboration between humans and robots in several applications. The only major concern with these supervised learning methods is that the effectiveness of these methods desires an ample amount of detailed labeled training data. Despite the need for a large amount of data for training and human supervision for labeling these data, making use of a robust alternative for supervised learning algorithms is a difficult task to accomplish in human action recognition.

 \begin{figure*}[tp]
 \makebox[\textwidth]{%
   \includegraphics[width=0.7\paperwidth]{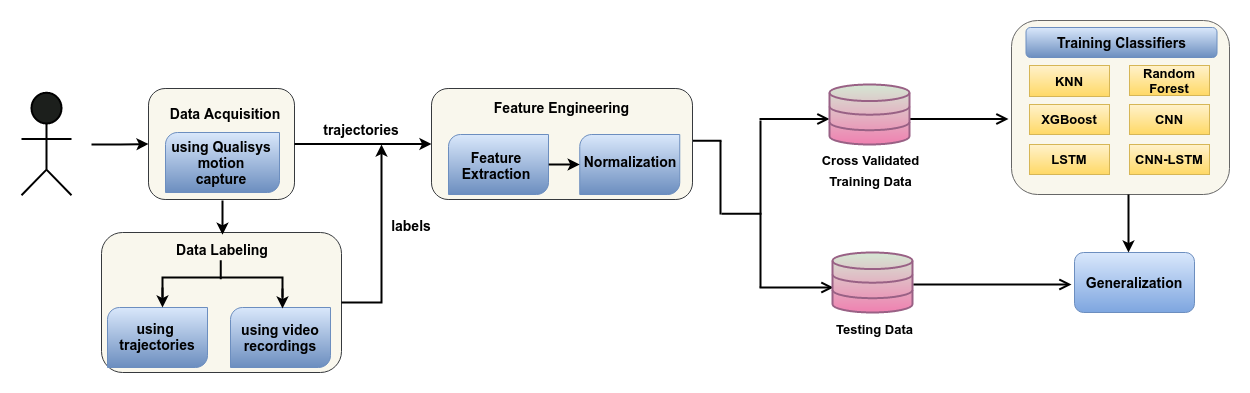}}
 \caption{Pipeline for creation of a human action recognition dataset and classification approach. Inspired from \cite{8736849}}
 \label{fig:flowchart}
\end{figure*}

The tasks concerning the action classification generally have four major phases: data acquisition, segment labeling, feature engineering, and finally training the classifier. The data can be acquired using different sensing modalities (for example video streams, IMUs, point clouds, etc.). If a sequence of several movements is recorded, the data needs to be preprocessed and segmented into
smaller movement entities and action labels need to be assigned to each segment.
Then, features have to be extracted from raw motion data and normalized.
Lastly, a classifier to recognize and infer the actions needs to be trained.
The pipeline for the creation of the dataset and labeling strategies used in this work is depicted in Figure \ref{fig:flowchart}.
The data labeling phase is a tedious and time-consuming process and poses a major constraint in the creation of a robust action recognition dataset. The data recorded using RGB cameras and RGB-D cameras are easy to obtain, and they provide rich appearance information. Thus, it makes the labeling task less complicated. On the other hand, sensor data recorded either using IMUs or marker-based motion capture system requires careful analysis of time series data to extract the stream of motions and assign a set of actions to it. There are previous works that propose to facilitate the data annotation process for time series data, e.g. Schröder et al. developed a tool support that makes use of a database schema for annotating sensor data~\cite{labelling_tool}. Cruciani et al. proposed a heuristic function based on step count and GPS information for online supervised training~\cite{s18072203}.
Another technique is to utilize few-shot learning methods as mentioned in~\cite{pub11335}, where small entities of human manipulation movements can be detected at high accuracy with $ \leq 10$ examples per class in the training data. Using the models trained on such a small dataset, it can be generalized to new unlabeled data.

In this paper, the influence of different labeling methods on the classification of human movements is investigated. Human movement is recorded based on a simple stacking scenario using a marker-based motion tracking system that measures the 3D positions of the human arm. 
Additionally, videos of the movements are recorded. 
After that, the recordings are labeled using two different methods. In the first labeling approach, the stacking movements are manually segmented using the video data. In the second method, recorded movement trajectories are automatically segmented into manipulation building blocks characterized by a bell-shaped velocity profile of the hand~\cite{pub7319} followed by manual correction of wrongly segmented data. The stacking movements are labeled carefully by examining the trajectory of the arm while performing the experiments using a labeling tool developed in-house, which visualizes the movement trajectories in 2D and 3D.  
Six different algorithms that are widely used for human action recognition, the k-Nearest Neighbor (KNN) classifier, Decision-Tree based classifier Random Forest (RF), Extreme Gradient Boosting (XGBoost), Deep Learning algorithms such as Convolutional Neural Networks (CNN), Long Short-Term Memory networks (LSTM), and a combination of CNN-LSTM networks are compared with respect to their suitability to label the movements automatically. The models are trained and evaluated on movements recorded at different speeds in order to study the influence of labeling techniques on the feasibility of transfer of speed in simple action movements. 
This paper is organized as follows: In section 2, an overview of related work is given. In section 3, the feature extraction and algorithms used for classification are described, along with the evaluation approaches. 
The data recording and the labeling procedure along with results and discussions are presented in
section 4 and section 5 respectively. The paper concludes with future scope of this work in section 6.

\section{\uppercase{Related Work}}
There is already a lot of work done on action recognition, involving vision-based and sensor-based modalities. The video streams provide rich spatial information and when it combines with the temporal information, i.e., image frames at time steps, it can be beneficial for the identification of human actions. Starting from the analysis of video streams, earlier works involved the usage of handcrafted feature-based methods such as the position of skeleton joints for action recognition tasks~\cite{6751553}.
In recent times, CNN-based approaches are quite popular because of their benchmarked results in computer vision problems and their ability to extract high-level representation in deep layers.
Donahue et al. introduced the Long-term Recurrent Convolutional Network (LRCN) consisting of a 2D CNN and LSTM for extracting RGB features and predicting action labels from each image ~\cite{inproceedings}. 
Ji et al. tried to capture the motion information from several adjacent frames by extracting spatial and temporal dimension features by performing 3D convolutions on the videos~\cite{6165309}.
The vision-based modalities require the use of an appropriately placed camera that poses mobility issues and privacy risks. With the availability of low-cost sensors and activity trackers in smartphones, a sudden shift has been observed in the usage of cameras for action recognition tasks.
Halloran et al. presented a comparison of Deep Learning models in human activity recognition on the MHEALTH dataset recorded using a smartphone~\cite{OHalloran2019ACO}.
They compared machine learning models on sensor-based data utilizing supervised learning methods.
Some recent works provide an alternative of using a labeled dataset and propose to train the model using semi-supervised methods, such as label propagation, requiring a less labeled dataset. 
Cruciani et al. proposed a heuristic function-based method for automatic labeling in an online supervised training approach~\cite{s18072203}. The algorithm generated weak labels by combining step count and GPS information.
Shamsipour et al. addressed the issue of labeling videos by considering only a few frames depicting the information of humans performing a particular activity~\cite{shamsipour2017human}. They randomly selected three video frames instead of employing all the frames, and used CNN for extracting features and SVM for classifying actions from the conceptual features.
However, the majority of the approaches in the literature are applied to whole-body human movements, such as walking, running, or sitting. In this work, we compared the classifiers' performance on movement building blocks that can be found in natural and intuitively performed movements and that can potentially be transferred to a robotic system using learning from demonstration~\cite{pub10504}. To our knowledge, there has been no previous work available that analyzes the influence of labeling techniques in movement classification and examines the possibility of speed transfer in action movements.

\section{\uppercase{Methods}}
In this section, the features that are extracted from the raw movement trajectories captured using a Qualisys motion capture system are described, along with the classification algorithms and the hyperparameter optimization methods used for training those algorithms.

\subsection{Feature Extraction}
Feature extraction is an important procedure in training machine learning algorithms. A meaningful representation of raw data can have a huge influence on the performance of predictive models. In this work, features are extracted from raw motion data in the same manner as mentioned in \cite{pub11335}.
The data is recorded using a marker-based motion capture system with several markers placed on the arm of the subject as shown in Fig \ref{fig:expSetup}. 
The marker positions are transformed into a global coordinate frame with respect to the markers placed on the back of the subject.
The features extracted directly from the raw data are the marker's 3D positions, velocity, orientation, and joint angle between them. The feature trajectories were interpolated to the same length and normalized in the range [0, 1].

\subsection{Classification Models}
\label{section:classificationModels}

\subsubsection{K-Nearest Neighbor}
KNN uses the proximity between the test data and already available training data to classify a sample. The closest proximity of the test data from the training data can be determined using distance metrics such as Euclidean, Manhattan, or Minkowski. We use KNN for comparison in this work because it performed exceptionally well on classifying human movement in a small-sized training dataset~\cite{Gutzeit2019}.
The feature trajectories for each movement recording are transformed into a 1-D feature vector. The closest neighbor of each data sample is determined using Euclidean distance, and the number of neighbors K required to classify the test data is tuned using grid search.

\subsubsection{Random Forest}
Random Forest is a bagging algorithm where random bootstrap samples are drawn from the training data and multiple decision trees are constructed. Each individual tree in the random forest outputs a class prediction, and the overall prediction results of the model are obtained by a voting approach on the individual decision tree outputs~\cite{ho1995random}. 
Due to the random selection of training data and features, the constructed decision trees are independent of each other, which makes the model resistant to overfitting problems. This improves its predictive performance and generalization abilities on unseen data. As the decision trees are generated in parallel at the time of training, it results in a speed-up of the training process. In this work, we have used a Random Forest classifier with grid search to tune the hyperparameters, such as the number of trees and the maximum depth of each decision tree in the algorithm. 

\subsubsection{Extreme Gradient Boosting}
XGBoost~\cite{Chen:2016:XST:2939672.2939785} is one of the most popular machine learning algorithms in recent times, widely used in structured and tabular data. XGBoost is a decision-tree-based ensemble algorithm that utilizes a gradient boosting framework and efficiently makes use of parallel processing and cache optimization for better speed and performance.
Boosting algorithms attempt to accurately predict the target variable by 
aggregating weak classifiers, which in general do not perform so well individually but when combined perform even better than the strongest individual learner. In order to make the final prediction, XGBoost sequentially adds the classifiers and fits the new model on the residual or the errors of the previous predictions that are then combined with previous trees to make the final prediction. It employs gradient descent when introducing new models to minimize the loss.

\subsubsection{Convolutional Neural Network}
In the last half of the decade, CNN has been one of the most popular variants of Neural Networks because of their substantial contribution and benchmarked results for computer vision, natural language processing~\cite{kim2014convolutional} and time series based problems \cite{2019}.
Contrary to images, CNNs in time series can be seen as a kernel sliding in only one dimension instead of two dimensions. In this work, a very simple Convolutional Neural Network is proposed consisting of two 1D convolution layers followed by a pooling layer, dense layer and an output layer with a softmax activation. The number of filters in the convolution layer, number of neurons in the dense layer, learning rate, and type of optimizer are modulated using Keras Tuner as mentioned in section \ref{section:kerasTuner}.

\subsubsection{Long Short-term Memory}
LSTMs were introduced by~\cite{hochreiter1997long} and are explicitly designed to avoid the long-term dependency problem exhibited in Recurrent Neural Networks. In Recurrent Neural Networks, there are directed cycles between the units, i.e., they propagate data forward and also backward and have the ability to process arbitrary long sequences of inputs using their internal memory. But where the problems of gradient explosion and gradient disappearance arise in RNN, LSTMs are able to avoid these problems with their architecture modifications, such as cell state, and its various gates. These gates are responsible to keep the essential information or forget it during training.
In this work, we use a simple structure with two LSTM
layers followed by a dropout layer and output layer with a softmax activation. 

\subsubsection{CNN-LSTM Network}
CNNs are known to be a robust feature extractor and capable of creating informative representations of time series data. On the other hand, LSTM networks perform pretty well at extracting patterns for long input sequences.
The combination of both the networks has shown good results on challenging sequential datasets~\cite{9065078}.
A CNN-LSTM model is used comprising two 1D convolution layers with ReLu activation function followed by 1D max-pooling layer and a flattening layer for formatting the feature map so that it can be consumed by the LSTM layer. 
Afterwards, the flattened feature maps from the previous layers are fed as an input to an LSTM layer followed by a dense layer and output layer with a softmax activation. We have also used a dropout layer, where randomly selected neurons are ignored during training. It helps in making the network capable of better generalization and less likely to overfit the training data. The number of layers, learning rate, and type of optimizer is modulated using Keras Tuner as mentioned in section \ref{section:kerasTuner}.
All the mentioned Deep Learning models are trained using sparse categorical cross-entropy loss function for 20 epochs using Adam optimizer and early stopping is used to stop the training if the accuracy on a validation dataset did not increase in the last $n$ epochs, where $n$ is called patience value to prevent the model from overfitting the training data. The experiments were automatically tracked using Weights \& Biases tool~\cite{wandb}.

\subsubsection {Hyperparameter Optimization }
\label{section:kerasTuner}
Hyperparameter tuning plays a vital role in the performance of machine learning algorithms. Hyperparameters can have a significant impact on model training in terms of model accuracy, training time, and computational requirements.
In this work, Keras Tuner~\cite{omalley2019kerastuner} that can be seamlessly integrated with Tensorflow 2 is used for tuning different hyperparameters required in training deep learning models. Methods used for tuning hyperparameters require defining a search space consisting of hyperparameters and their ranges that are needed to be optimized. Some hyperparameters that are optimized in this work are number of filters in convolution layers, number of units in LSTM layer, number of neurons in the dense layer and, choice of optimizers.
%

For each of the above-mentioned parameters, a range of its possible values is provided. Keras Tuner supports different search heuristics (such as random search, Hyperband, and Bayesian optimization) that helps in finding the best value of the defined hyperparameter to enhance the model performance.
In this work, a Bayesian optimization search strategy is utilized for the optimization of hyperparameters. It eliminates the problem of choosing a combination of different hyperparameters randomly that could sometimes result in the abysmal combinations of parameters. Thus, resulting in failure to improve the model accuracy.  
Instead of choosing random combinations, the Bayesian optimization strategy chooses the best possible hyperparameters based on the model performance of previous combinations. It constructs a probabilistic representation of the performance of a given Machine Learning algorithm, which is modeled using Bayesian inference and Gaussian process.

\subsection{Evaluation Approach}
\label{section:evaluationApproach}
The six algorithms described in section \ref{section:classificationModels} are compared 
on the movement recordings labeled using two different methods as mentioned in section \ref{section:experiments}. The classifiers are trained on the data comprising arm movements performed by the subject at a slow and normal speed (section \ref{section:experiments}) and evaluated on the movements performed by the subject at a fast speed. The main focus of the experiments was to investigate the impact of labeling techniques on the generalization ability of the model for speed transfer in the fast-paced movement. These movements were appropriate for evaluating the trained model because segmentation of the fast-paced movements into basic action movements is more challenging compared to slower-paced movements and requires precise tracking of the subject's arm position.
The training data was divided into 5 folds using stratified cross-validation. In each split, the evaluation is performed on the unseen test data. Finally, generalization accuracy with a standard deviation of mean accuracy and F1 score is reported. The dataset was completely balanced with an equal number of data for each class.

\section{\uppercase{Experimental Data}}

\subsection{Stacking Scenario Data}
\label{section:experiments}

\begin{figure}[th]
\centering
\includegraphics[width=0.8\columnwidth]{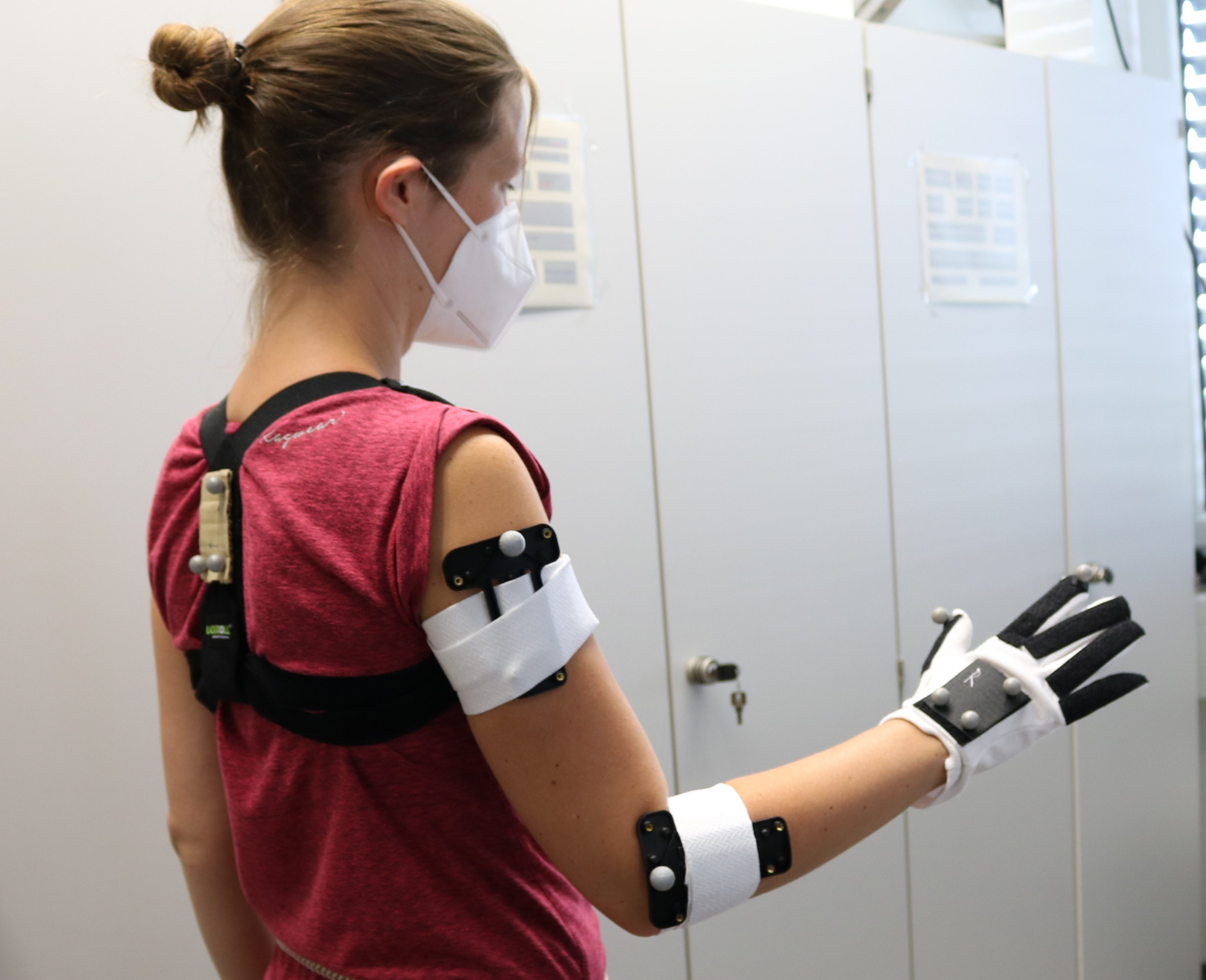}
\caption{Stacking-scenario setup. Positions of markers attached on the arm and the back of the subject are recorded
using a camera based motion tracking system.}
\label{fig:expSetup}
\end{figure}

\begin{figure}[th]
\centering
\includegraphics[width=0.9\columnwidth]{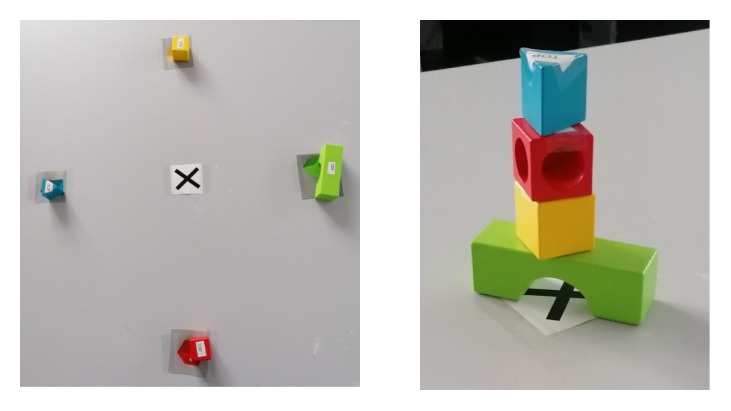}
\caption{Stacking-scenario setup. The left image shows the different positions of the bricks on the table, and the right image shows one of the stacking examples. The bricks were stacked at the position of the cross.}
\label{fig:Stacking}
\end{figure}

The experiment was conducted on a single subject and movements were recorded with a Qualisys motion tracking system that uses infrared light reflecting markers. Additionally, the performed movements were recorded using a video camera. Markers were attached to the right hand, elbow, shoulder and back of the subjects, as shown in Figure~\ref{fig:expSetup}. The marker positions were tracked with 7 Qualisys cameras and data was recorded at 500Hz. The subject was asked to perform a basic stacking movement as shown in Figure~\ref{fig:Stacking} where bricks of different colors were placed on fixed positions on the table and the participant was asked to place the bricks in the middle of the table by stacking it one by one. The experiment was performed by arranging the bricks in different stacking order: the green brick was always kept at the bottom while the other bricks (red, blue, yellow) were arranged in different permutations. Thus, overall, 6 different stacking orders were recorded with three repetitions of each stacking order.
The movement for stacking the bricks was recorded at three different speeds (slow, normal, fast) for all 6 stacking orders. The normal and slow speed were intended to provide a comfortable speed for placing bricks from their respective position to the middle of the table one over another, while the fast speed challenged the participant. There were many instances where the bricks were not successfully placed over one another due to the fast arm movement, resulting in bricks tripping over the table.

\subsection{Labeling Techniques}
The movement data was decomposed using two different ways into 8 classes (middle2front, front2middle, middle2left, left2middle, middle2right, right2middle, middle2down, down2middle) based on the position from where the bricks were supposed to be picked and placed.
In the first labeling technique, it was segmented using the video recorded for the experiment that was synced precisely with the data recorded from the Qualisys motion tracker. The labeling method was quite tedious and time-consuming, as one has to cautiously track the image frames on the video where the subject picks and places the bricks. Although tracking the movement of the data recorded at slow and normal pace was quite precise, labeling the data recorded at fast pace was very challenging.
In the second labeling technique, the arm movement data was automatically segmented using a velocity-based probabilistic segmentation presented in~\cite{pub7319} into basic movement units with a bell-shaped velocity. The unnecessary trajectories were removed, and essential segments required for training the model were annotated using a labeling tool developed at our institute, which visualizes the movement trajectories in 2D and 3D. Using this tool, inaccurate segment boundaries of the automatic segmentation approach were corrected.
After labeling the data using both methods, the dataset consisted of $144$ arm movements each for slow and normal paced recordings and $192$ arm movements for fast-paced recordings, that means in total 480 labeled movement sequences were available.

\subsection{Complexity of the Dataset}
\begin{figure}[th]
\centering
\includegraphics[width=0.9\columnwidth]{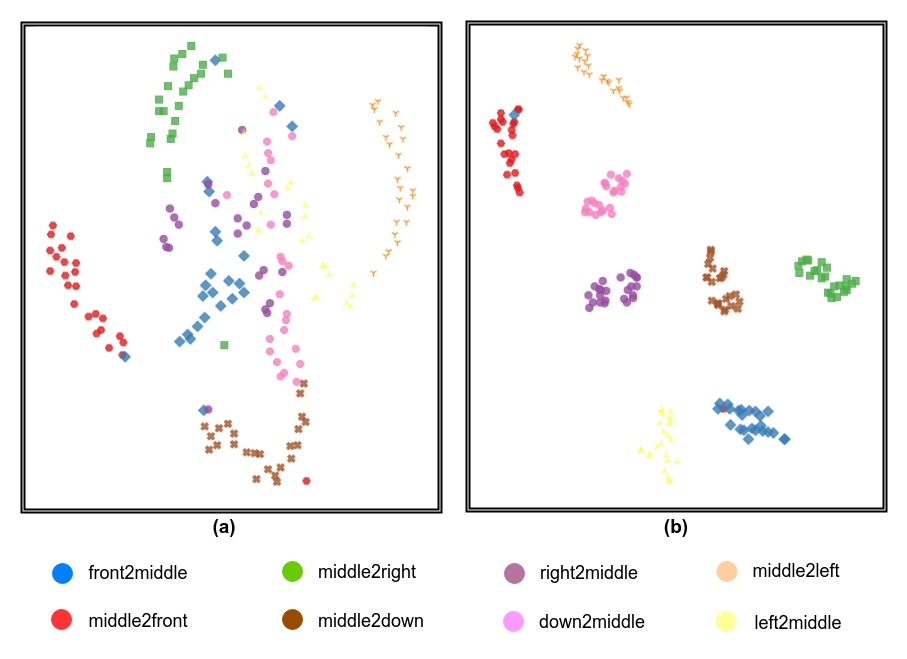}
 \caption{T-SNE plots of the stacking scenarios recorded at a fast pace (a) Data labeled using Videos  (b). Data labeled using movement trajectories. Each action movement labels can be identified by a different color.}
 \label{fig:TSNE}
\end{figure}

In this section, we compare the structure and diversity of the movement recordings labeled using above-mentioned techniques. For understanding the overlapping between different classes in the dataset, t-SNE (t-distributed Stochastic Neighbor Embedding) introduced by~\cite{vanDerMaaten2008} is used for exploring a set of points in a high-dimensional space by transforming it into a lower-dimensional space. In Figure~\ref{fig:TSNE}, we can see that data from the stacking scenarios labeled using the movement trajectories is less complex and are separated more clearly, but one can see overlap in at least 4 classes in the segments labelled using videos.

\section{\uppercase{Results and Discussions}}
\begin{figure*}[th] 
\centering
\subfloat[]{\includegraphics[width=0.85\columnwidth]{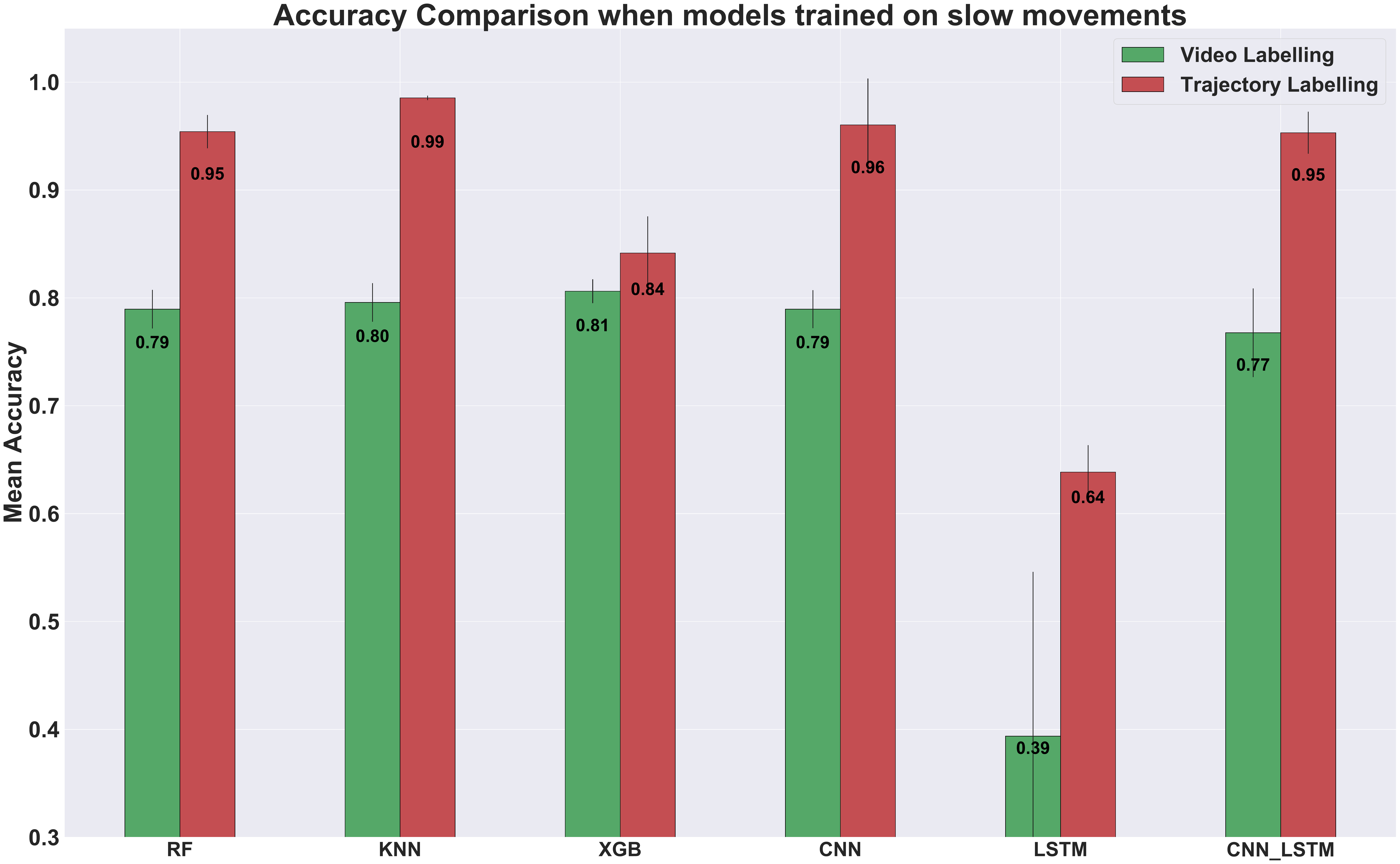}}
\qquad
\subfloat[]{\includegraphics[width=0.85\columnwidth]{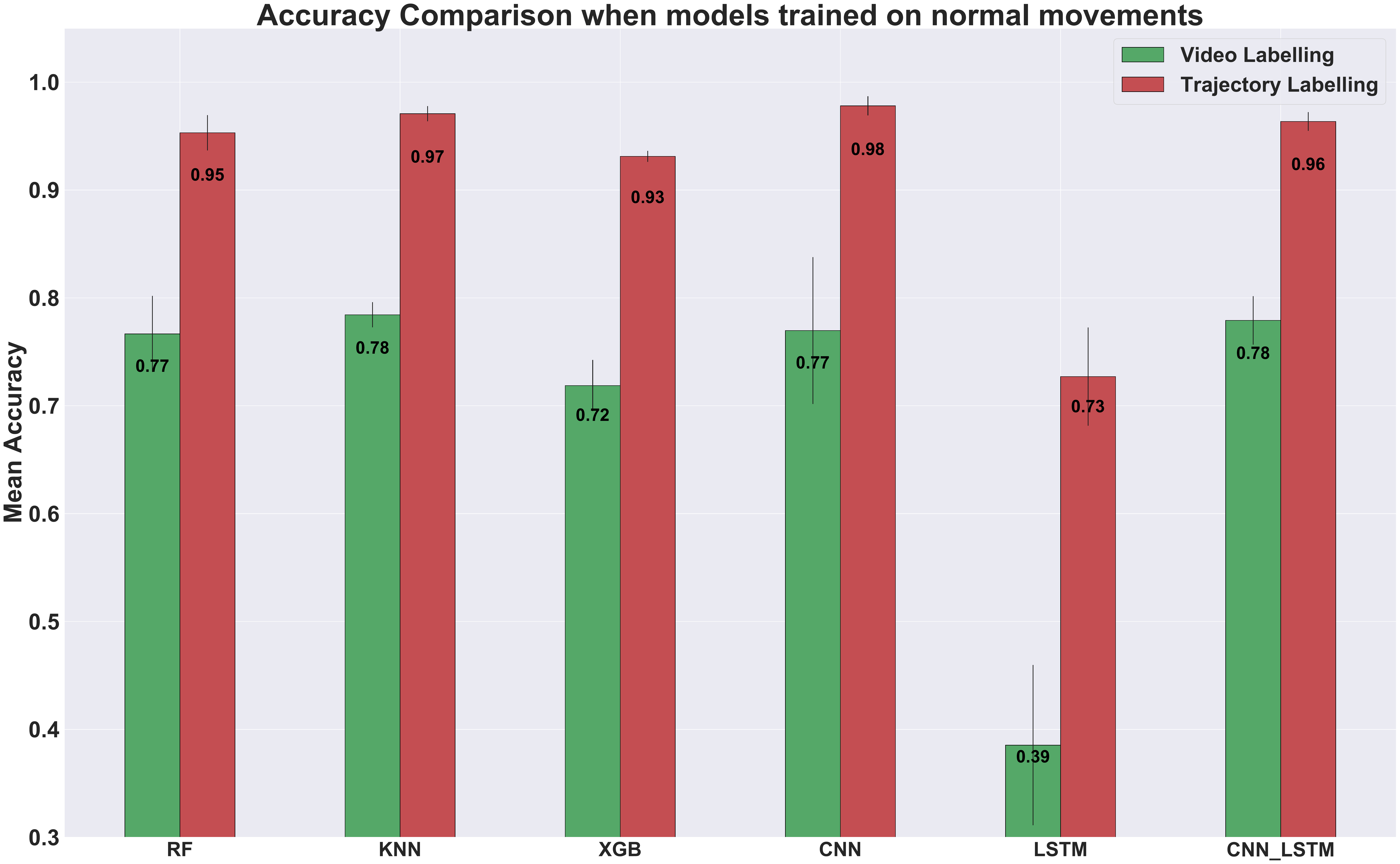}}
\qquad
\subfloat[]{\includegraphics[width=0.85\columnwidth]{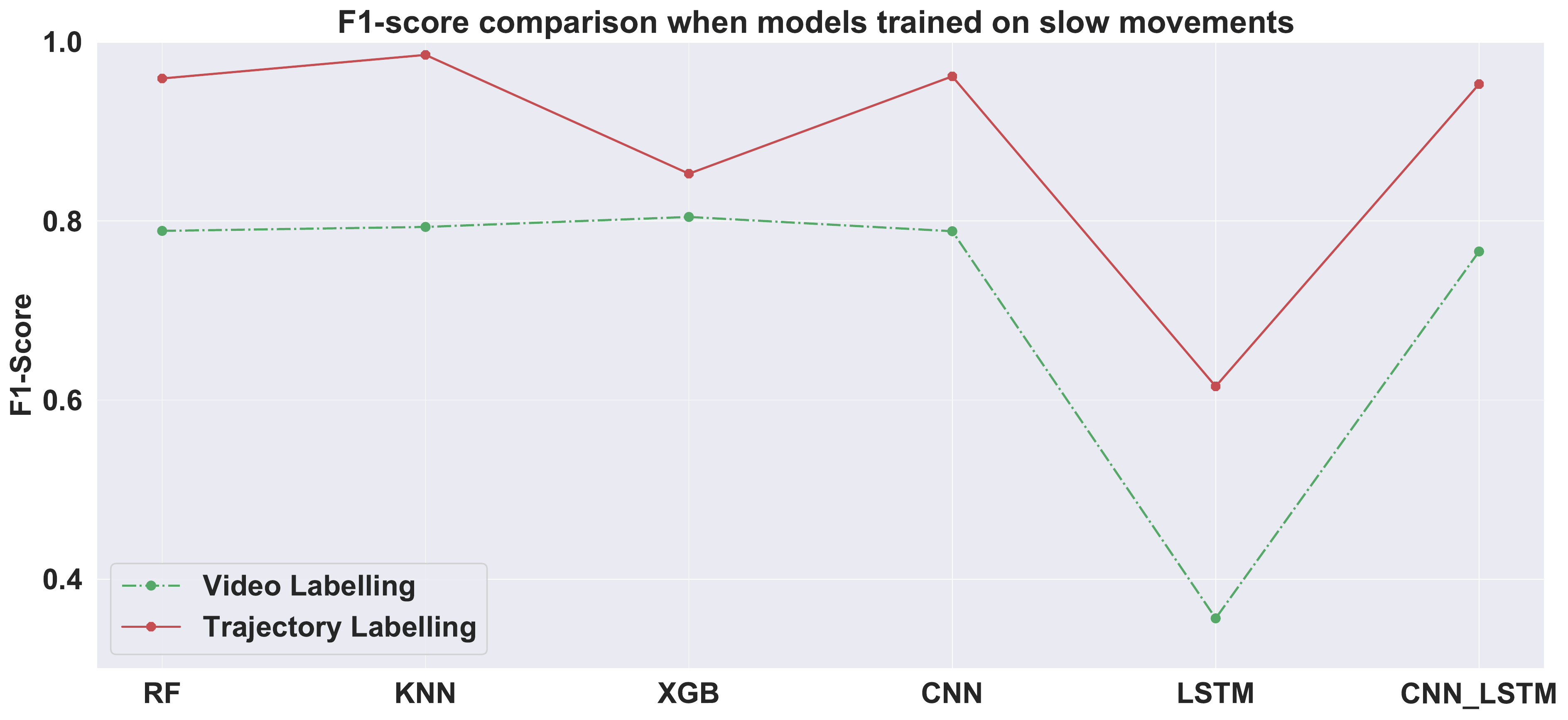}}
\qquad
\subfloat[]{\includegraphics[width=0.85\columnwidth]{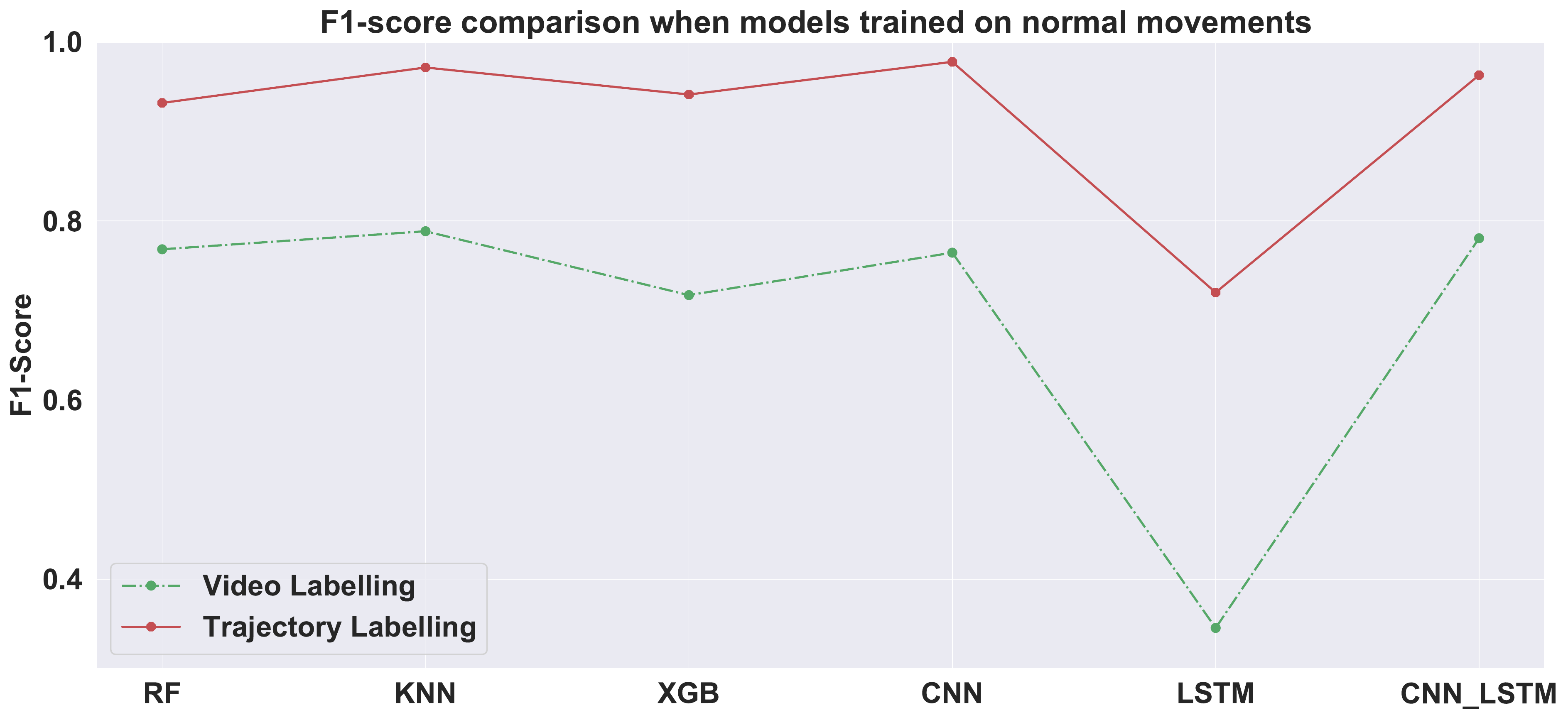}}

\caption{(a) and (c) Accuracy and F1 score comparison when model trained on slow movements. (b) and (d) Accuracy and F1 Score comparison when model trained on normal movements.}

\label{fig:results}
\end{figure*}

In this work, the generalization ability of different models, such as KNN, Random Forest, XGBoost, CNN, LSTM, and CNN-LSTM model, to data recorded at different speeds are compared using two different strategies.
The main aim was to demonstrate the influence of labeling methods in movement speed transfer within human movements. The classifiers are trained on data consisting of movements recorded at a slow and normal pace and tested on data recorded on fast-paced movements.
The results of the generalization capabilities of the classifiers are shown in Figure~\ref{fig:results}.
As we can see from the plots, all the classifiers have a much better generalization on the fast-paced movements when training data is labeled using movement trajectories.
The results illustrate the significance of labeling strategies and their impact on classification accuracy irrespective of the choice of classifiers. 
For the model trained on normal movements and labeled using trajectories, there is almost 20\% improvement in accuracy for all the models except LSTM that has an accuracy difference of approximately 35\%. CNNs are the best performing model with a mean accuracy of  98\% and F1-score of 97.7\% for normal movements labeled using trajectories.
For the model trained on slow movements and labeled using trajectories, KNN was the best performing model with an accuracy of 99\% and an F1-score of 98.5\%, and
all models except XGBoost have an accuracy difference of approximately equal to or greater than 16\%. Precise labeling of the recordings from videos requires meticulous tracking of the arm by the labeling person, and there are chances of human errors in tracking the accurate frames from the videos. That could be the reason for low accuracy on data labeled using videos.
As the dataset consists of movements derived from simple stacking scenarios, distance-based algorithms performed considerably better and were almost equivalent to CNNs in performance. The LSTM classifiers fail to generalize well, but providing more data by employing augmentation techniques and training for more epochs could further enhance the results.


\section{\uppercase{Conclusion and Future Work}}
In this paper, we studied the impact of annotation quality on the classification accuracy on data consisting of basic human movements. Two different labeling strategies have been proposed and six different Machine Learning and Deep Learning models were compared. The potential possibility of speed transfer using the model trained on the data labeled using these two strategies was examined. It is found that fast-paced movements are better recognized on data labeled using trajectories of the recorded movements. The best results
could be achieved with k-Nearest Neighbor and CNNs, achieving an accuracy of 99\% and 98\% on the model trained on slow and normal paced movements respectively. \\
For future work regarding the movement recognition models, some self-supervision methods that showed promising results in the field of Computer vision and NLP domain could be explored and there are possibilities to leverage such networks for sensor data in human action recognition. Although it does not completely discard the usage of labeled data, it learns useful representations of the data from the unlabeled dataset, which can then be fine-tuned on a small number of labeled data. 
Further, more focus could be given to traditional machine learning methods for model explainability. It would help to prevent model bias and could help to understand the working of a model in a better way. Shapely values \cite{DBLP:journals/corr/LundbergL17} and LIME \cite{DBLP:journals/corr/RibeiroSG16} can give a rough idea about the features that greatly influenced the performance of the classifiers. Model interpretability has a huge prospect in the AI community, one can compare the models used in this work based on their interpretability to get a better understanding of these black-box models for human action recognition tasks.\\
Regarding the influence of the labeling techniques, the experiments conducted in this paper were performed
with just one subject on simple movement data. For a deeper investigation of this influence, the movements of more subjects and more complex movements should be analyzed.
Furthermore, not only the labeling technique but also the experience of the person labeling the data should be taken into account.
However, the first small study presented in this paper already shows that accurately segmented data could significantly improve the movement classification accuracy.

\vspace{0.5cm}
\section*{\uppercase{Acknowledgements}}
This work was supported through a grant of the German Federal Ministry for
Economic Affairs and Energy (BMWi, FKZ 50 RA 2023).

\vspace{-0.5cm}

\bibliographystyle{apalike}
\bibliography{literature}
%

\end{document}